\documentclass[10pt,twocolumn,letterpaper]{article}

\usepackage{iccv}
\usepackage{times}
\usepackage{epsfig}
\usepackage{graphicx}
\usepackage{amsmath}
\usepackage{amssymb}

\usepackage[margin=0.92in]{geometry}
\usepackage{caption}
\usepackage{subcaption}

\usepackage[pagebackref=true,breaklinks=true,letterpaper=true,colorlinks,bookmarks=false]{hyperref}

\usepackage[capitalize]{cleveref}
\crefname{section}{Sec.}{Secs.}
\Crefname{section}{Section}{Sections}
\Crefname{table}{Table}{Tables}
\crefname{table}{Tab.}{Tabs.}

\iccvfinalcopy % *** Uncomment this line for the final submission

\begin{document}

%%%%%%%%% TITLE
\title{Blendshapes GHUM: Real-time Monocular Facial Blendshape Prediction}

\author{
Ivan Grishchenko \quad Geng Yan \quad Eduard Gabriel Bazavan \quad Andrei Zanfir \quad Nikolai Chinaev\\
Karthik Raveendran \quad Matthias Grundmann \quad Cristian Sminchisescu\\
Google\\
1600 Amphitheatre Pkwy, Mountain View, CA 94043, USA\\
{\tt\footnotesize \{igrishchenko, gengyan, egbazavan, andreiz, chinaev, krav, grundman, sminchisescu\}@google.com}\\
}

\date{}

\maketitle
% Remove page # from the first page of camera-ready.
% \ificcvfinal\thispagestyle{empty}\fi

%%%%%%%%% ABSTRACT
\begin{abstract}
We  present  Blendshapes GHUM  –  an  on-device  ML pipeline that predicts 52 facial blendshape coefficients at 30+ FPS on modern mobile phones, from a single monocular RGB image and enables facial motion capture applications like virtual avatars. Our main contributions are: i) an annotation-free offline method for obtaining blendshape coefficients from real-world human scans, ii) a lightweight real-time model that predicts blendshape coefficients based on facial landmarks.

\end{abstract}

%%%%%%%%% BODY TEXT
\section{Introduction}

Facial motion capture is widely used in movies and video games to create expressive characters that are driven by human actors. It has been used more recently in video chats and live streaming where a user can puppeteer a virtual avatar using their face. Early methods employed marker dots to track landmarks on the face \cite{10.1145/97879.97906}, while other approaches have removed this requirement by incorporating multiple cameras \cite{10.1145/1198555.1198596}, sensors \cite{bradley2010high}, and data-driven techniques\cite{Zhang2008}. In this paper, we introduce a marker-less, real-time facial motion capture pipeline that only uses the RGB camera input on devices like mobile phones and laptops.

There have been numerous approaches to facial animation over the last two decades ranging from physics-based rigs \cite{sifakis}, parametric face models \cite{10.1007/978-4-431-66890-9_1}, PCA bases \cite{Zhang2008}, blendshapes \cite{10.2312:egst.20141042}, to generative approaches (GANs) \cite{DBLP:journals/corr/abs-1906-06337}. We focus on blendshapes, which are semantically meaningful and artist-controllable representations of facial expressions. In a typical workflow, technical artists begin with a neutral 3D mesh of the character to be animated. Then, they pick a set of basic expressions (\eg smile, frown, eyes closed) and create corresponding posed meshes (blendshapes) in a 3D modeling program. Any expression can be reconstructed by computing the weighted sum of these blendshape displacements from the neutral mesh. While they do not form a strictly orthogonal basis, blendshapes are linear, relatively sparse, and an interpretable means of representing human expressions \cite{10.2312:egst.20141042}. Unlike many other methods, blendshapes do not require the target to resemble a human face and offer a great deal of artistic freedom to design expressive characters. In our setup, we use the same set of 52 blendshapes as those in Apple’s ARKit \cite{ARKit}, that are familiar to many 3D modeling studios and animators. 

\begin{figure}
  \begin{center}
   \includegraphics[width=0.9\linewidth]{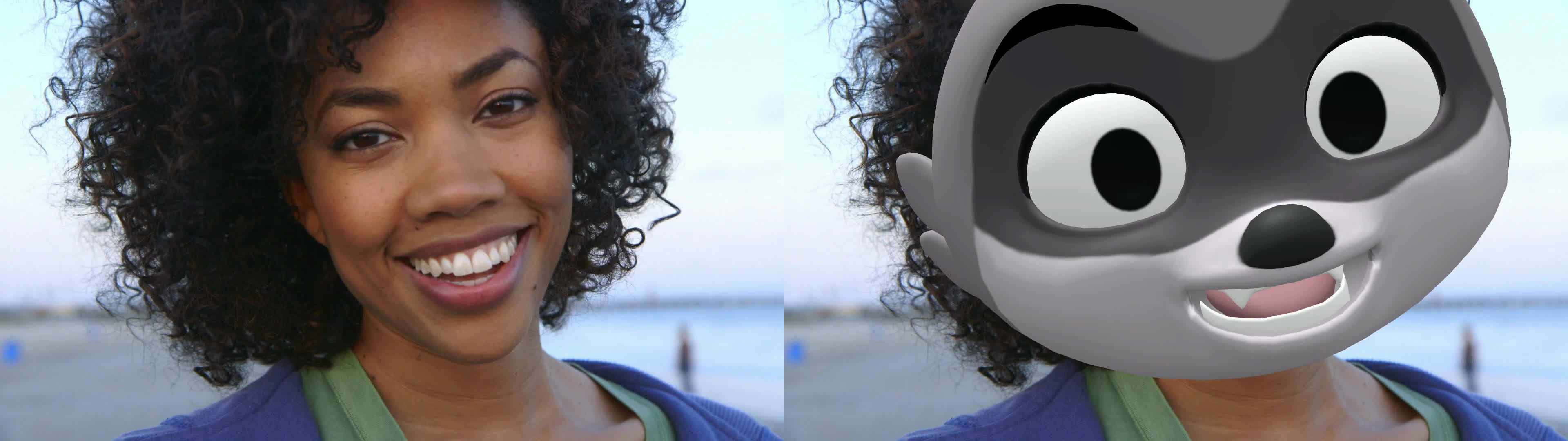}
   \includegraphics[width=0.9\linewidth]{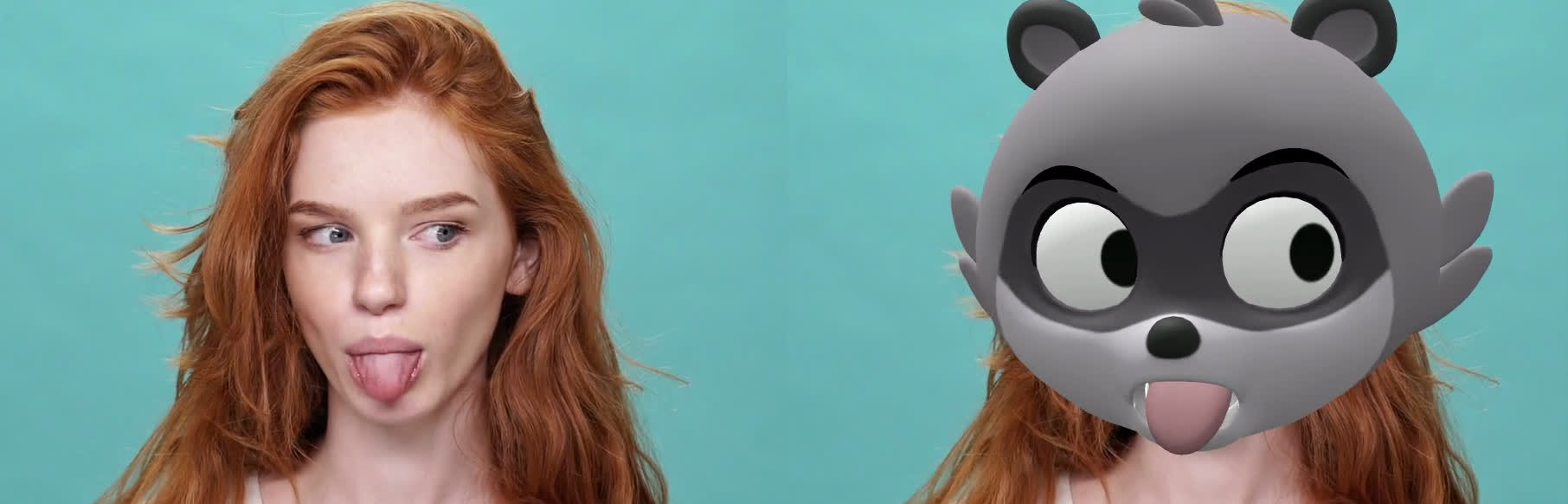}
  \end{center}
  \caption{End-to-end example of our blendshape pipeline which takes in a single RGB image and predicts blendshape coefficients to animate a virtual avatar to match the expression of the person in the image.}
  \label{fig:blendshapes_example}
\end{figure}

In order to animate a virtual character using these blendshapes, we need to compute the corresponding set of coefficients from the provided input. We introduce a calibration-free technique that only requires a single RGB image as input by using deep neural networks to detect and predict facial landmarks and subsequently, blendshape coefficients from these landmarks (\cref{fig:blendshapes_example}). Our paper is structured as follows:

\begin{figure*}[t]
  \begin{center}
   \includegraphics[width=1.0\linewidth]{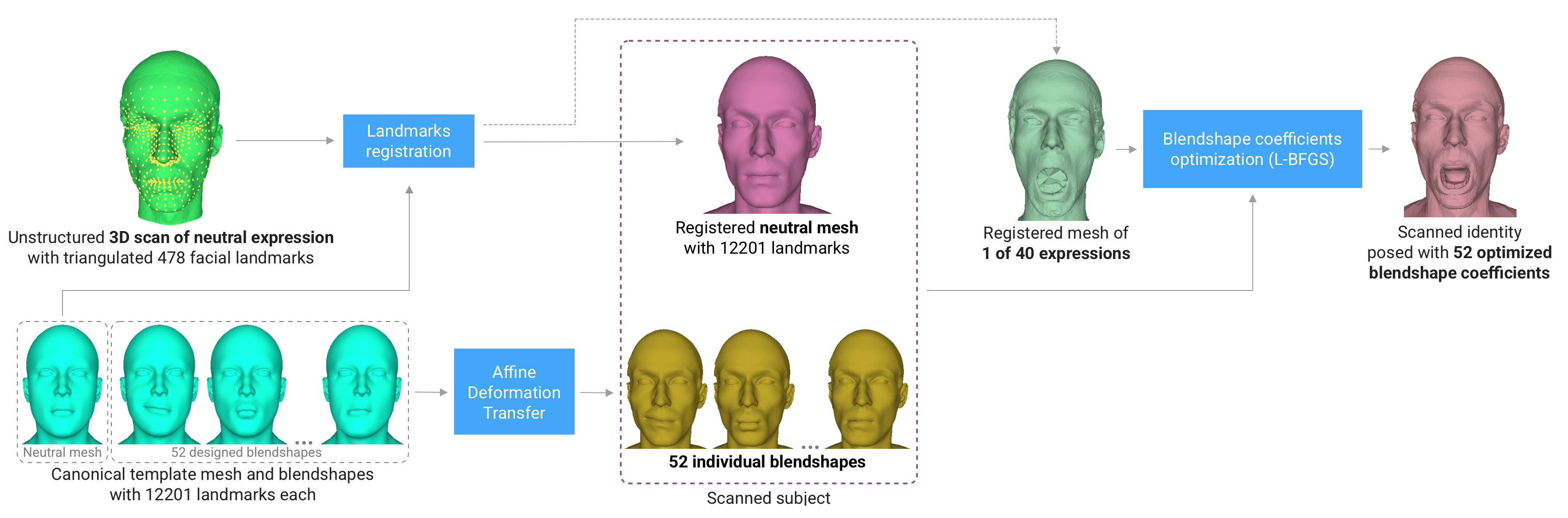}
  \end{center}
  \caption{Overview of the offline blendshapes registration process.}
  %\textbf{Landmarks registration.} From the lab we get unstructured 3D scans of subjects and 478 3D landmarks obtained by multi-view triangulation of the face mesh model predictions from \ref{facial_landmarks_section}. We use canonical template mesh produced by an artist and register it to unstructured scans optimizing to keep facial contours from 478 landmarks and mesh semantics from the template, and produce semantically consistent meshes with 12201 3D landmarks. The one with neutral expression is taken as neutral mesh. \textbf{Blendshapes transfer.} We use 52 blendshapes designed by an artist on canonical mesh and transfer them to neutral mesh of the scanned subject using affine deformation. \textbf{Blendshape coefficients optimization.} For each subject we recorded 2 second clips of them performing 40 expressions. We then optimize blendshape coefficients using L-BFGS method so that when individual subject blendshapes are mixed, they align with registered landmarks for each scan from the recorded sequence.}
  \label{fig:offline_registration_schema}
\end{figure*}

\begin{itemize}
\item We describe a convolutional neural network that can accurately predict a set of facial landmarks for a wide range of human expressions.
\item We introduce an offline data acquisition and fitting pipeline that uses registered 3D scans of human subjects to produce neutral shape, blendshapes and blendshape coefficients.
\item We train a real-time neural network on this data to predict blendshape coefficients purely from facial landmarks and evaluate its performance.
\end{itemize}

%------------------------------------------------------------------------
\section{Facial landmarks} \label{facial_landmarks_section}

We use facial landmarks as the basis for our blendshapes pipeline. We collected a set of $33,792$ images covering a wide range of human expressions. Each image is labeled with $478$ 2D facial landmarks\cite{mediapipefacemesh} by a team of human annotators. This is a relatively easy task for the annotators, compared to directly annotating blendshape coefficients corresponding to a particular expression. We use this annotated data as ground truth to train a lightweight model based on MobileNetV2\cite{sandler2018mobilenetv2}.
%We use the topology of the publicly available MediaPipe face mesh [], %since it has a high density of landmarks along salient contours on eyes, %yebrows, irises and lips, and this allows us to capture small changes in %facial expressions.

The model takes in $256\times256$ images, produces $128\times$ downsampled feature maps via inverted residual blocks, and then uses a fully connected layer to predict the coordinates of $478$ landmarks. We trained this model on  $8\times8$ TPUv3\cite{CloudTPU}  with batch size of $4,096$ and an initial learning rate of $0.001$. We term the mean L2 distance between predicted and ground-truth landmarks normalized by the inter-ocular distance as Mean Normalized Error (MNE) and use it as the basis for comparisons in the rest of the paper. We achieved an MNE of $2.71$\%   (\cref{realtime-model-evaluation}) on our test set which compares favorably to the human annotation error of $2.56$\% (computed via cross-validation between $11$ annotators on a set of $58$ images).  This model runs at 8 ms on a Pixel 6 phone via the Tensorflow Lite (TF-Lite) OpenCL inference backend and is a good fit for real-time usage.

% \begin{figure}[t]
%   \begin{center}
%   \begin{subfigure}[t]{0.2\textwidth}
%     \centering
%     \includegraphics[height=0.8in]{iccv2023AuthorKit/fig/lip_attn.png}
%     \caption{Attention Mesh}
%     \end{subfigure}%
%     ~
%     \begin{subfigure}[t]{0.2\textwidth}
%     \centering
%     \includegraphics[height=0.8in]{iccv2023AuthorKit/fig/lip_v2.png}
%     \caption{New Face Mesh}
%     \end{subfigure}
%   \end{center}
%   \caption{Improved Face Mesh accuracy on lips. The new Face Mesh is able to capture the expressions better, like the subtle smiles above.}
%   \label{fig:facemesh_improvement_example}
% \end{figure}

% \begin{table}[!h]
% \begin{center}
% \begin{tabular}{| c |c |c |c |} 
% \hline
%  & \textbf{All} & \textbf{Eyes} &	\textbf{Lips} \\
% \hline
% \textbf{Proposed Face Mesh} & $\mathbf{2.71}$ & $\mathbf{3.56}$ &	$\mathbf{2.42}$  \\
% \hline
% Annotators cross-validation & $2.56$ & 2.15 & 2.51 \\
% \hline
% \end{tabular}
% \caption{\label{mae-facemesh}Proposed Face Mesh model average per-landmark 2D error normalized by interocular distance compared with AttentionMesh}
% \end{center}
% \end{table}

\paragraph{Tracking accuracy} To confirm that our facial landmarks model can capture all 52 blendshape activations, we performed a qualitative visual user study that consisted of 10 videos per blendshape (activating from neutral to the maximum), for a total of 520 videos. We rendered the predicted face mesh on each video and asked a set of 3 annotators to assess if the mesh captured the expression being enacted by the human, and picked the majority opinion per video. We found that $96$\% of the videos successfully met this criteria, which indicates that it can serve as an accurate approximation of facial expressions. 

\begin{figure*}[htb]
  \begin{center}
   \includegraphics[width=0.9\linewidth]{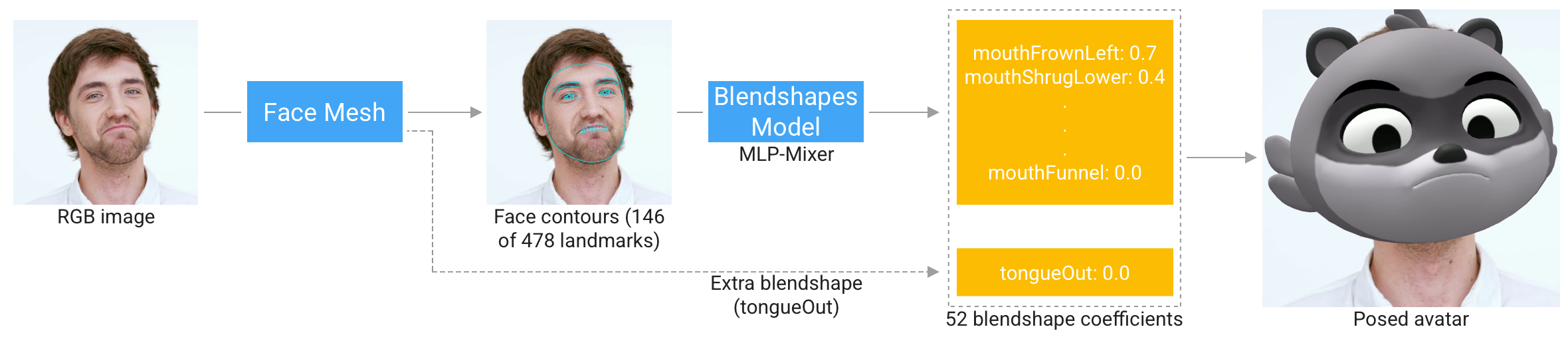}
  \end{center}
  \caption{Overview of the Blendshapes GHUM inference pipeline.}
  %From a single RGB input image we predict 478 2D facial landmarks. A lightweight MLP-Mixer model predicts 51 blendshapes from lips, eye and eyebrow contours. Tongue state (int/out) is predicted directly by the Face Mesh model as landmarks have no information about it. The avatar is posed with total of 52 blendshape coefficients.
  \label{fig:inference_pipeline}
\end{figure*}

%------------------------------------------------------------------------
\section{Data acquisition for blendshapes}

\subsection{Animation with blendshapes} In order to animate a 3D character using blendshapes, we need a neutral mesh ($\mathbf{b_{0}}$) corresponding to the neutral expression of the character, and meshes ($\mathbf{b_i}$) for each fully-activated expression (\eg a full smile, or eyes fully open) that we term as blendshapes. To reconstruct a specific expression, we apply the weighted sum of  blendshape displacements to the neutral mesh as seen in \cref{eq:blendshapes_sum}, where $\mathbf{b_i}$ denotes the mesh corresponding to the $i$-th blendshape and $w_i$ denotes its coefficient. Note that only semantically meaningful combinations of blendshapes are expected to yield plausible results.

\begin{equation}
    \label{eq:blendshapes_sum}
    \mathbf{b} = \mathbf{b_0} + \sum_{n=1}^{52} w_i (\mathbf{b_i} - \mathbf{b_0})
\end{equation}

We focus on the inverse problem in this paper and solve for the neutral mesh, blendshapes and blendshape coefficients from a single image of a human in real-time, so that these can be used to drive a virtual character.

\subsection{Offline data processing}
We first solve an offline version of this problem in a calibrated setup, where we have access to multi-camera recordings of people performing various expressions. We then generate data to train a real-time model that approximates this process.

\paragraph{Lab scans} Our proprietary lab setup allows us to capture detailed scans of human heads at 60Hz. Each scan results in a 3D mesh of the head using RGB images captured from fixed cameras. We scanned $6,000$ different people performing $40$ predefined expressions. Expressions were selected in a way that they are easy to perform by an untrained human and that each of the $52$ blendshapes was used in at least one expression. Each expression recording is a two second clip that starts with a neutral expression and ends with the maximum activation of that expression. 

\paragraph{Landmarks template registration} Template registration transforms unstructured 3D scans into a unified representation with consistent topology and semantics. Our canonical template mesh contains $12,201$ vertices compared to the $468$ from the face mesh for improved registration. We ran our face landmark model on each frame of video and projected the outputs onto the surface of the corresponding scan to obtain 3D landmarks for registration. We used a version of the technique described in \cite{ghum2020}, based on \textit{as-conformal-as-possible} surface priors \cite{yoshiyasu2014conformal}, but guided by much denser and accurate 3D landmarks. % We retrained the variational autoencoder to use a spherical embedding and ensured that the facial contours of the template aligned with those on the 3D scan.  This alignment improves the accuracy of the entire registration. This registration aligns each 3D scan with the template mesh and establishes an identity-specific neutral mesh per subject.

\paragraph{Optimizing blendshapes and coefficients} We created 52 blendshapes for the canonical template mesh with the help of a technical artist. We then transferred these blendshapes to each scanned identity in their neutral pose that we obtained from the previous step via affine deformation transfer \cite{sumner2004deformation, li2010example}. We thus obtain a set of identity specific blendshapes for each scanned subject. 

Then, we find optimal blendshape coefficients to generate a mesh that aligns with each frame of the scanned subjects enacting various expressions. We use a quasi-Newton method (L-BFGS \cite{liu1989limited}) to optimize the blendshape coefficients by introducing slack variables to softly constrain them to be between $0$ and $1$. 
%The objective is the Euclidean distance between the blended 3D landmarks and the pseudo-ground-truth landmarks available for each scan. 
We also add a global rigid transformation \cite{zhou2018continuity} to better align the result. Finally, we clip the coefficients to remain between $0$ and $1$. As a result of this process, we have facial landmarks as well as blendshape coefficients corresponding to various expressions across a range of identities that we can use as the basis for training a real-time model.

\paragraph{Blendshapes registration accuracy} We analyzed the accuracy of our registration process by comparing the landmarks on the scanned mesh posed with optimized blendshape coefficients to the landmarks predicted by the face mesh model from the frontal view of the scan. Since the face mesh model only predicts landmarks in 2D, we project the 3D mesh of the posed face to 2D using known camera parameters. We obtained a MNE of $3.67\%$ which indicates that the offline registration is capable of capturing human facial expressions accurately (\cref{realtime-model-evaluation}).

%------------------------------------------------------------------------
\section{Real-time blendshapes model}\label{realtime_model_section}

We train a lightweight model that takes 2D face contours produced by the face mesh model as input and produces 52 blendshape coefficients as output. We generate training data by sampling from scanned 3D face models and predefined combinations of blendshapes.

\paragraph{Training data} While our scanned data covers a large number of identities, it limit us to $40$ recorded expressions per person. We expanded our training set by generating more expressions for each identity. We first selected a random 3D face model from the $6000$ scanned ones. Next, we sampled random blendshape coefficients based on our constructed prior described below and apply them to the selected neutral identity mesh to produce a unique expression corresponding to that subject. Finally, we applied a random 3D transformation and projected it to 2D using known camera parameters. As a result, we obtained 2M pairs of facial landmarks and  corresponding ground truth blendshape coefficients which serves as our training data. We restrict ourselves to 146 landmarks on the lips, eyes, eyebrows, irises and face oval contours as they are most crucial for expression tracking.

\paragraph{Blendshape coefficients prior}
Not all combinations of blendshapes yield plausible human expressions. We avoided creating non-realistic training examples and expanded on the range of expressions by constructing a prior for sampling blendshape coefficients. We grouped semantically similar blendshapes into basic expression groups and defined possible combinations between them. For example, "smile left" and "smile right" were grouped together and could be activated either symmetrically or independently. Next, we created expression groups by face region (\eg mouth, eye, eyebrow, iris) and allowed independent activations in each group. We allowed the activation of several basic expressions at the same time for certain regions to allow for more subtle expressions: \eg the mouth region has about 15 basic expressions and we allowed activations of up to $4$ at once. Finally we resolved some possible conflicts: \eg 
the coefficient for "mouth close" can't be greater than that of "mouth open". We further refined this prior via visual and numerical evaluation.

\paragraph{Model} Our real-time model is based on a lightweight MLP-Mixer architecture \cite{DBLP:journals/corr/abs-2105-01601}. It takes $146\times2$ tokens as input, converts them into a $96\times64$ latent representation, and finally predicts 52 blendshape coefficients and a 6D facial rotation matrix via two separate 2D convolutions.

%via 2D convolutions and performs 4 iterations of "mixing". Each "mixing" includes consequent $\times4$ expansion and squeezing along each side of the latent tensor. 
The training loss consists of two parts: an L2 loss for predicted blendshape coefficients and an L2 loss for landmarks of the mesh reconstructed by applying predicted and ground truth blendshape coefficients to the sample identity. The first loss ensures fidelity to the ground truth coefficients while the second enforces perceptual accuracy and takes into account non-orthogonality of the blendshapes basis.

We train the model with a batch size of 512 for 50K steps, gradually reducing the learning rate from $10^{-3}$ to $10^{-5}$. The resulting model runs at 1.2ms on Pixel 6 phone via TF-Lite XNNPACK inference backend\cite{xnnpack}.

%\paragraph{Face Mesh adjustment} As the generated facial landmarks (based on the template) used in training can differ slightly from those predicted by the face mesh during inference, we add a small amount ($1$mm) of noise to bridge this domain gap. This noise is applied to the posed 3D face model before projecting it to 2D and was determined empirically to improve the robustness of the model while allowing it to retain the ability to track subtle changes in expression.

%------------------------------------------------------------------------
\section{Evaluation}

Our real-time pipeline outlined in \cref{realtime_model_section} takes in an RGB image, predicts facial landmarks and then passes them to the blendshapes model to produce $52$ blendshape coefficients. Since blendshape bases are non-orthogonal, multiple combinations of blendshapes can yield very similar expressions. This many-to-one mapping makes direct comparisons of blendshape coefficients error-prone and noisy for a given expression.

Instead, we opted to quantitatively evaluate the real-time blendshapes model via facial landmarks. We used a holdout set of 500 face scans, reconstructed meshes and corresponding blendshapes. We then posed these neutral meshes with blendshape coefficients predicted by the real-time blendshapes model. We finally projected these 3D meshes to 2D for that view using known camera parameters from the calibrated lab setup. We ran the face mesh model on the frontal view images from these scans to obtain 2D landmarks that served as ground truth.

We computed the MNE between corresponding pairs of 146 landmarks on the set of contours described in \cref{realtime_model_section}.

\paragraph{Baselines} 

Evaluation results are reported in \cref{realtime-model-evaluation} with additional metrics on lips and eyes. The real-time blendshapes model has an overall MNE of $3.88\%$.

The cross-validation MNE of $2.33\%$ on facial contours from the human annotators serves as an estimate of the lower bound on the error one might expect from the model. We also computed this metric between pairs of fully activated blendshapes on the canonical template mesh and artist designed blendshapes. These serve as an upper bound on the errors and provide a measure of the differences between two expressions (\eg how much "mouth smile" differs from "mouth frown"). We obtain a mean $8.45\%$ error across all pairs of blendshapes divided into lips and eye regions \cref{realtime-model-evaluation}.

The accuracy of our blendshapes model is much closer to the achievable human accuracy than it is to the upper baseline. This, in combination with visual evaluation, gives us confidence in the quality of the proposed approach for blendshape prediction.

Please refer to the supplementary material for a qualitative evaluation of the model on unseen video sequences. 
%Each recording includes original video stream, facial landmarks produced by the face mesh model and an avatar animated with predicted blendshape coefficients.

\begin{table}[!h]
\begin{center}
\begin{tabular}{| c |c |c |c |} 
\hline
 & Lips & Eyes & Average \\
\hline
Annotators cross-validation & $2.51$ & $2.15$ & $2.33$  \\
\hline
Face Mesh model & $2.42$ & $2.39$ & $2.41$  \\
\hline
Offline registration & $2.67$ & $4.31$ & $3.67$  \\
\hline
Pairwise blendshapes diff & $8.50$ & $8.40$ & $8.45$  \\
\hline
\textbf{Real-time model} & $\mathbf{3.76}$ & $\mathbf{3.95}$ & $\mathbf{3.88}$  \\
\hline
\end{tabular}
\caption{\label{realtime-model-evaluation} Comparison of MNE (in \%) on eyes and lip regions.}
\end{center}
\end{table}

%------------------------------------------------------------------------
\section{Conclusion}

We present a pipeline for the real-time prediction of blendshape coefficients from a single RGB image by using face landmarks as the intermediate representation. We showed that these landmarks serve as an excellent proxy for capturing human expressions and that the resulting blendshapes model generalizes across identities and expressions. 

We will provide details about our open-sourced models upon publication. Future work could include the explicit prediction of the neutral mesh and blendshapes basis through the blendshapes model apart from the coefficients.

%\paragraph{Open source} We open-sourced our real-time blendshapes model and posted it alongside with various demos and APIs on mediapipe.dev. Web demo can be found at XXX. We received warm feedback from metaverse community with our solution now being using in major platforms like VTuber Studio and Three.js.

%\paragraph{Future work} Potentially impactful improvement would be to include identity prediction in real-time blendshapes model. It will allow to accumulate knowledge about the user's neutral shape and blendshapes over time and thus improve blendshape coefficients prediction (vs. having it done implicitly by the blendshapes model now).

% Another good improvement would be to further optimize transferred blendshapes during offline blendshape coefficients fitting. It will blendshapes basis even more individual and capable to capture all user expressions.

{\small
\bibliographystyle{ieee_fullname}
\bibliography{egpaper_for_review}
}

\end{document}